\documentclass[letterpaper, 10 pt, conference]{ieeeconf}  

\IEEEoverridecommandlockouts                              

\usepackage{graphics} 
\usepackage{epsfig} 
\usepackage{mathptmx} 
\usepackage{times} 
\usepackage{amsmath} 
\usepackage{amssymb}  

\usepackage{algorithm}
\usepackage{algorithmicx}
\usepackage[noend]{algpseudocode}
\usepackage{array}
\usepackage{textcomp}
\usepackage{verbatim}
\usepackage{booktabs}
\usepackage{multirow}
\usepackage{makecell}
\usepackage{xcolor}

\usepackage[table]{xcolor}
\usepackage{cite}
\usepackage{hyperref}

\DeclareMathAlphabet{\mathcal}{OMS}{cmsy}{m}{n}

\title{\LARGE \bf
MA3DSG: Multi-Agent 3D Scene Graph Generation \\ for Large-Scale Indoor Environments
}

\author{}
\author{Yirum Kim$^{1}$, Jaewoo Kim$^{1}$, Ue-Hwan Kim$^{1}$$^\dagger$
\thanks{$^{1}$ All authors are with the Department of AI Convergence, Gwangju Institute of Science and Technology (GIST), Gwangju 61005, Republic of Korea. {\tt\small \{kimyirum,kjw01124\}@gm.gist.ac.kr}}
\thanks{$^\dagger$ Corresponding author: Ue-Hwan Kim {\tt\small uehwan@gist.ac.kr}}
}

\begin{document}

\maketitle
\thispagestyle{empty}
\pagestyle{empty}

\begin{abstract}
Current 3D scene graph generation (3DSGG) approaches heavily rely on a single-agent assumption and small-scale environments, exhibiting limited scalability to real-world scenarios. In this work, we introduce Multi-Agent 3D Scene Graph Generation (MA3DSG) model, the first framework designed to tackle this scalability challenge using multiple agents. We develop a training-free graph alignment algorithm that efficiently merges partial query graphs from individual agents into a unified global scene graph. Leveraging extensive analysis and empirical insights, our approach enables conventional single-agent systems to operate collaboratively without requiring any learnable parameters. To rigorously evaluate 3DSGG performance, we propose MA3DSG-Bench—a benchmark that supports diverse agent configurations, domain sizes, and environmental conditions—providing a more general and extensible evaluation framework. This work lays a solid foundation for scalable, multi-agent 3DSGG research. 
\end{abstract}

\section{Introduction} \label{sec:intro}
3D scene graph generation (3DSGG) serves as a cornerstone for comprehensive high-level 3D scene understanding. By detecting objects and describing their relationships via predicates, it provides valuable context for diverse tasks such as image captioning \cite{Wang2020LearningVR, Chen2020SayAY}, image generation \cite{Yang2022DiffusionBasedSG}, change detection \cite{Looper20223DVL}, navigation \cite{Li2022RemoteON, Gadre2022ContinuousSR}, and task planning \cite{Agia2022TaskographyER, Rana2023SayPlanGL}.

Since the introduction of the 3DSGG benchmark \cite{Armeni20193DSG}, extensive research has primarily focused on improving \textit{performance}—through enhanced relational reasoning \cite{Lv2023SGFormerSG}, the integration of structured prior knowledge \cite{Zhang2021Knowledgeinspired3S, Wang2023VLSATVS, Feng20233DSM}, and the adoption of open-vocabulary settings \cite{Koch2024Open3DSGO3, Chen2024CLIPDrivenO3}---while relatively little focus has been given to the \textit{scalability} of the methods. As machine agents are increasingly deployed across a diverse and expanding set of real-world domains \cite{Vincent2022AnIF, Okafuji2021BehavioralAO}, the ability to not only achieve strong performance but also sustain it at scale \cite{Busch2023EnablingTD, Cao2024DeepRL} has become a necessity. This raises a fundamental question: ``\textbf{\textit{Are current 3DSGG methods scalable?}}"

\begin{figure}[t]
  \centering
  \includegraphics[width=\linewidth]{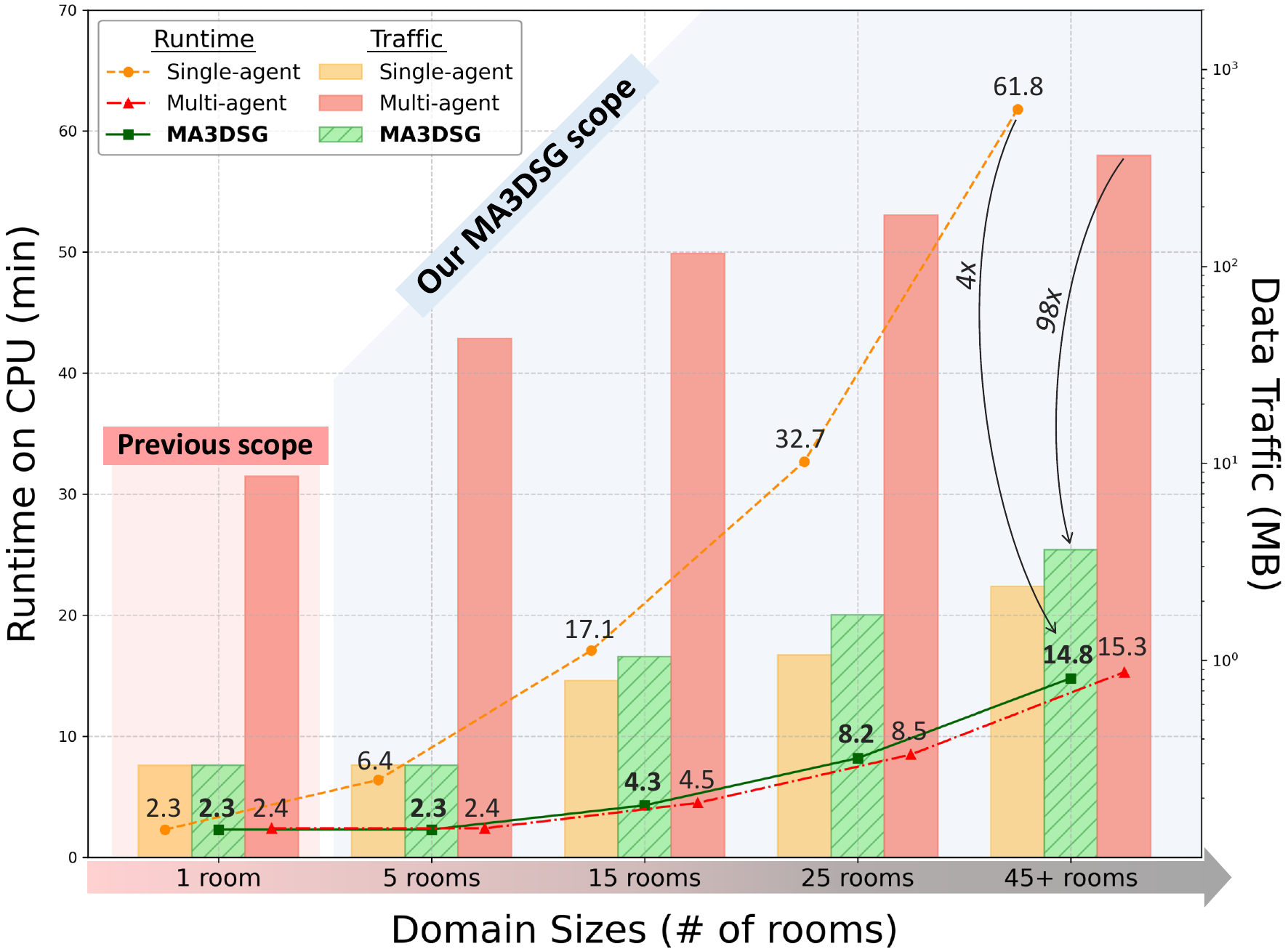}
  \vspace{-5mm}
  \caption{Comparison of Runtime and Data Traffic. Our MA3DSG (\textit{14.8 min, 3.7 MB}) runs $4\times$ faster than single-agent system (SGFN, \textit{61.8 min}), and uses $98\times$ less data traffic than multi-agent system (SGFN + SG-PGM, \textit{364.1 MB}) in extremely large-scale environments. Unlike the single-agent baselines and MA3DSG, which were only executed on CPUs, the multi-agent baselines utilized GPUs on the backend due to their model complexity.}
  \label{fig:bandwidth}
  \vspace{-5mm}
\end{figure}

Our findings, as illustrated in Figure~\ref{fig:bandwidth}, demonstrate that contemporary 3DSGG methods encounter significant scalability challenges---with runtimes up to \textit{\textbf{4× longer}} (vs. single-agent methods) and data traffic up to \textit{\textbf{98× heavier}} (vs. naive multi-agent methods) compared to our proposed MA3DSG. We attribute these scalability issues to two key limitations: (1) the prevalent reliance on single-agent paradigms; and (2) benchmarks biased toward constrained environments.

Lastly, to enable rigorous evaluation of both performance and scalability, we introduce \textbf{\textit{MA3DSG-Bench}}---a flexible and extensible benchmark that supports diverse agent configurations (single- and multi-agent), varying scales (ranging from 1 to 47 rooms), and scene dynamics (static and long-term condition). In contrast to prior 3DSGG benchmarks \cite{Wang2023VLSATVS, Koch2024Open3DSGO3} that process each room in 3RScan \cite{Wald2019RIO3O} independently using a single agent, our benchmark treats all reference rooms as a unified navigable space, enabling parallel exploration by multiple agents. Furthermore, we incorporate rescan sequences \cite{Wald2019RIO3O} to reflect realistic long-term environmental changes. By facilitating joint perception and temporal context modeling in dynamic multi-agent environments, our MA3DSG-Bench sets a new standard in the field.

To summarize, our contributions are as follows:
\begin{enumerate}
    \item \textbf{Problem Formulation.} We extend the 3DSGG task to multi-agent, large-scale settings. To the best of our knowledge, this is the first holistic effort to address the scalability challenges in 3DSGG research.

    \item \textbf{Model Design.} We propose MA3DSG built for the 3D semantic scene graph domain---featuring an efficient graph alignment algorithm. MA3DSG demonstrates strong scalability across diverse domain sizes while ensuring fast and robust performance.

    \item \textbf{Benchmark Setup.} We introduce a comprehensive MA3DSG-Bench that expands previous single-agent, small-scale-only 3DSGG benchmarks. This contribution provides a solid foundation to guide and inspire future research in scalable 3DSGG.
\end{enumerate}

\section{Related Work}
\subsection{3D Scene Graph Generation}
Current research on 3DSGG can broadly fall into two areas based on their perspectives on 3D scene representations: hierarchical 3D scene graphs and 3D semantic scene graphs---the focus of our work.

\subsubsection{Hierarchical 3D Scene Graphs} Hierarchical 3D scene graphs organize entities such as buildings, rooms, objects, and cameras into a unified structure \cite{Armeni20193DSG}. Several works have enlarged the estimation of such hierarchical 3D scene graphs to large-scale environments. For example, 3D dynamic scene graphs handle scenes with moving agents \cite{Rosinol20203DDS}, Kimera builds 3D dynamic scene graphs from visual-inertial data \cite{Rosinol2021KimeraFS}, and Hydra incrementally constructs the layers of a hierarchical scene graph \cite{Hughes2022HydraAR}. While these approaches~\cite{Armeni20193DSG, Rosinol20203DDS, Rosinol2021KimeraFS, Hughes2022HydraAR, Udugama2023MonohydraR3, Chang2023HydraMultiCO} provide a compact and efficient representation of 3D scene environments, they primarily indicate the existence of entities for expressing relationships rather than capturing their detailed semantic nuances of how those objects are configured and interact. As a result, they lack the expressive power required for diverse downstream tasks such as task planning \cite{Agia2022TaskographyER}, scene change detection \cite{Looper20223DVL}, and manipulation of 3D scenes \cite{Dhamo2021Graphto3DEG}.

\subsubsection{3D Semantic Scene Graphs}
In contrast to hierarchical 3D scene graphs, 3D semantic scene graphs focus on inter-object semantics and contextual interactions \cite{Kim20193DSG, Feng20233DSM}. Following the development of the 3D scene graph dataset \cite{Wald2020Learning3S} built on top of 3RScan \cite{Wald2019RIO3O}, 3D semantic scene graph generation from reconstructed point clouds has emerged: GNN-based analysis \cite{Zhang2021ExploitingER}, performance enhancement through prior knowledge \cite{Zhang2021Knowledgeinspired3S}, instance embedding based generation \cite{Wald2022Learning3S}, language-based contrastive pre-training \cite{Koch2023Lang3DSGLC}, and visual-linguistic semantics assisted training \cite{Wang2023VLSATVS}. Another stream of work has proposed to incrementally construct a 3D semantic scene graph from image sequences and depth data \cite{Wu2021SceneGraphFusionI3, Wu2023Incremental3S}.

By explicitly delineating spatial relationships between objects and their surroundings, these graphs enhance a variety of downstream applications, including 3D point registration \cite{sarkar2023sgaligner}, 3D scene reconstruction \cite{Dhamo2021Graphto3DEG}, change detection \cite{Looper20223DVL}, and task planning \cite{Agia2022TaskographyER}. Despite these advancements, conventional approaches hardly address the inherent scalability limitations of the 3D scene graph generation process. In this work, we concentrate on 3D semantic scene graph research and tackle the critical yet underexplored challenge---facilitating a \textit{scalable} 3D semantic scene graph generation for large-scale environments.

\begin{figure*}[!t]
  \centering
  \includegraphics[width=0.98\linewidth]{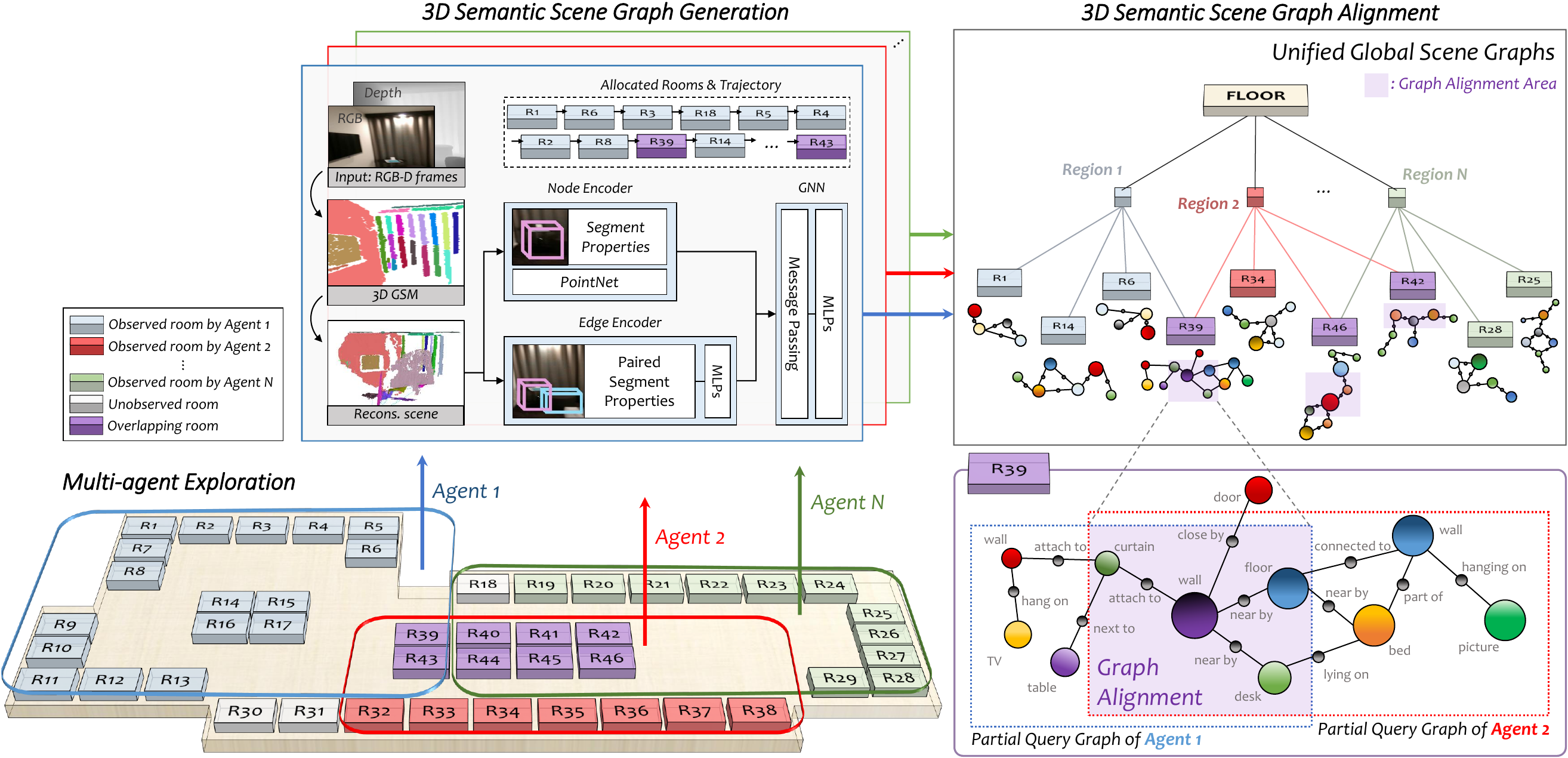}
  \caption{The overall architecture of the proposed MA3DSG. Each agent incrementally generates 3D semantic scene graphs in a large-scale environment. The framework consists of multi-agent exploration, 3D semantic scene graph generation, and graph alignment, where agents collaboratively construct and integrate local scene graphs into a unified global representation.}
  \label{fig:framework}
  \vspace{-4mm}
\end{figure*}

\subsection{Multi-Agent System}
Multi-agent systems have been extensively studied for their potential to enhance the robustness and scalability of single-agent frameworks by harnessing the synergistic capabilities of a swarm. In the context of simultaneous localization and mapping (SLAM), recent works \cite{Lajoie2023SwarmSLAMSD, Cramariuc2022AM} enable agents to explore collaboratively through exchanging sensor data and jointly optimizing 3D maps. Similarly, previous studies \cite{Zou2019CollaborativeVS, Chang2023HydraMultiCO} present multi-agent cooperation frameworks for fusing data from heterogeneous robots in hierarchical 3D scene graph generation. Furthermore, a diverse range of approaches, including multi-domain cooperation \cite{Yang2022MultiDomainCS}, probabilistic occupancy mapping \cite{Sunil2023FeatureBasedOM}, optimized cooperative exploration and communication \cite{Lajoie2023SwarmSLAMSD, Schmuck2017MultiUAVCM}, and distilled collaboration graphs \cite{Li2021LearningDC}, have also demonstrated the versatility of multi-agent systems. Building on these notable advancements, we extend multi-agent methodologies to the realm of a 3D semantic scene graph framework. To the best of our knowledge, our work represents the first comprehensive effort to develop a robust multi-agent 3DSGG system, accompanied by a detailed and rigorous benchmark.

\subsection{Graph Alignment}
Graph alignment aims to maximize structural and attribute consistency across graphs, and has traditionally been formulated as a graph isomorphism or quadratic assignment problem \cite{aflalo2015convex, Yan2016ASS, emmert2016fifty}. However, classical approaches often suffer from high complexity as the graph size increases \cite{aflalo2015convex, raveaux2016exact}, fail to capture edge structures \cite{Yu2018GeneralizingGM}, or require equal-sized graphs \cite{Maretic2019GOTAO}. Recently, deep learning-based methods \cite{wang2023deep,wang2024graph} learn flexible matching functions and robust node representations. In the domain of 3D scene understanding, partial graph matching using geometric and semantic features improves alignment robustness and downstream performance \cite{sarkar2023sgaligner,wang2024graph}. Nevertheless, current learning-based techniques are often limited by high training complexity, computational overhead, and slow inference speed, which restricts their applicability in large-scale or real-time scenarios.

\section{Methodology}

\subsection{Problem Formulation}
For each agent $k \in \{1, \ldots, K\}$, we define a collection of RGB-D observation sequences as $\mathcal{S}^k=\left\{ s_r^k \right\}_{r=1}^{R_k}$, where $R_k$ denotes the number of rooms visited by agent $k$. Each sequence $s_r^k$, which corresponds to the scanned data of the $r$-th room explored by agent $k$, is a temporally ordered sequence of frames as $s_r^k = \left( I_{r,1}^k, I_{r,2}^k, \ldots, I_{r,T^{k}_{r}}^k \right)$, where $I_{r,t}^k$ denotes the RGB-D frame captured at time step $t$ and $T^{k}_{r}$ is the total number of frames recorded in room $r$ by agent $k$. Based on these observations, the objective of MA3DSG is to let each agent incrementally construct the local 3D scene graph $\mathcal{G}^k_r = (\mathcal{V}^k_r, \mathcal{E}^k_r)$, where $\mathcal{V}^k_r$ denotes the set of semantic entities (nodes) and $\mathcal{E}^k_r$ denotes the relationships (edges) between entities for room $r$. These partial graphs are integrated into a unified global scene graph $\mathcal{G}'= \bigcup_{r=1}^{R}\mathcal{G}^k_r$ in a decentralized and collaborative manner. We address the challenges posed by dynamic scenes and the integration of conflicting partial graphs, using a function $f : G^{1}_r \times G^{2}_r \times \dots \times G^{n}_r \;\rightarrow\; \mathcal{G}'$ with $G^{k}_{r}$ defined in its own frame $\mathcal{F}^{k}_{r}$. Notably, the scene graph is updated over time as $\mathcal{V}_r(t+1) = \mathcal{V}_r(t) + \Delta \mathcal{V} $ and $ \mathcal{E}_r(t+1) =  \mathcal{E}_r(t) + \Delta \mathcal{E}$, thereby ensuring the adjustment to long-term dynamic objects of the environment.

\subsection{Overview}
We propose a decentralized multi-agent framework for 3D scene graph generation, addressing the need for scalable and efficient scene understanding in large-scale environments. Unlike single-agent approaches constrained to small domains, MA3DSG utilizes multiple agents to collaboratively explore diverse regions, as shown in Figure~\ref{fig:framework}. It consists of three core components: (1) \textit{Multi-Agent Exploration}, enabling distributed coverage of large spaces; (2) \textit{3D Semantic Scene Graph Generation}, where agents incrementally build local 3D semantic scene graphs; and (3) \textit{3D Semantic Scene Graph Alignment}, a lightweight algorithm that integrates local graphs into a unified global representation. By sharing information through overlapping exploration, MA3DSG enhances completeness and reduces overhead---increasing its utility in real-world scenarios.

\subsection{3D Semantic Scene Graph Generation}
\subsubsection{3D Global Segmentation Map (3D GSM)}
Each agent performs incremental geometric segmentation \cite{Tateno2017CNNSLAMRD} on the input RGB-D sequence to generate the 3D global segmentation map (GSM). 3D GSM consists of multiple segments $U=\{u_1, u_2,\dotsc, u_n\}$, with each segment containing the point cloud $P_i=\{\mathrm{p}_i \mid \mathrm{p}_i \in \mathbb{R}^3\}$. With each new incoming frame, 3D GSM is updated by incorporating new segments or eliminating old ones. Each segment $u_i$ is characterized by multiple properties: the centroid $\overline{\mathrm{p}}_i \in \mathbb{R}^3$, the standard deviation of the points $\mathrm{\sigma}_i$, the size of the axis-aligned bounding box $\mathrm{b}_i=(b_x,b_y,b_z) \in \mathbb{R}^3$, the maximum length $l_i=max\{b_x,b_y,b_z\}$, and the volume $\nu_i = b_x \cdot b_y \cdot b_z$. Subsequently, segments for each instance are integrated into the reconstructed scene, with each instance treated as a node.

\subsubsection{Feature Graph}
Leveraging the seminal SGFN \cite{Wu2021SceneGraphFusionI3}, the node encoder generates node features $v_i$ by extracting a latent feature vector $E(\mathrm{P}_i)$ of the point cloud using PointNet \cite{Qi2016PointNetDL}.  To address scale insensitivity from normalization, spatial-invariant properties are concatenated to $E(\mathrm{P}_i)$. The edge encoder, consisting of three multi-layer perceptrons (MLPs), computes edge features $e_{ij}$ for any two neighboring nodes $i$ and $j$ $(i \ne j)$ by processing their relative spatial properties. Formally, the node/edge features are defined as:
{
\begin{gather}
  v_i=[E(\mathrm{P}_i), \mathrm{\sigma}_i, ln(\mathrm{b}_i), ln(\nu_i), ln(l_i)], \\
  e_{ij} = f_s([\Delta\overline{\mathrm{p}}_{ij}, \Delta\mathrm{\sigma}_{ij}, \Delta \mathrm{b}_{ij}, \ln\left(\frac{\nu_i}{\nu_j}\right), \ln\left(\frac{l_i}{l_j}\right)]),
\end{gather}
}where [·] denotes a concatenation function and $f_s$ represents MLPs. The feature-wise attention network ($FAN$) \cite{Wu2021SceneGraphFusionI3} is employed, where node and edge features are robustly updated in the message passing layer as follows:
{\begin{gather}
    v_i^{m+1} = f_v\left(v_i^m, \max_{j \in \mathcal{N}(i)} \left( FAN(v_i^m, e_{ij}^m, v_j^m) \right)\right), \\
    e_{ij}^{m+1} = f_e\left(v_i^m, e_{ij}^m, v_j^m\right),
\end{gather}
} where $f_v$ and $f_e$ denote MLPs and $\mathcal{N}(i)$ denotes the set of neighbors of node $i$.

\subsection{3D Semantic Scene Graph Alignment}
Merging a stored 3D scene graph with a newly generated graph is a significant challenge due to the complexity of the subgraph matching process, which arises from variations in graph size and structural differences. The task is inherently NP-hard, and its complexity is further exacerbated by the dynamic nature of objects in our test benchmark scenarios, where positions and states are subject to change. To address this, we propose a novel incremental graph alignment algorithm that seamlessly integrates newly generated 3D scene graphs with existing data. The algorithm enhances robustness by aligning the new graphs with stored data, updating nodes and edges that have changed, and leveraging prior information to infer unscanned regions. The proposed graph merging consists of two key stages:
\begin{itemize}
    \item \textbf{Graph Alignment}: The agent identifies the intersection subgraph between the newly generated query graph and the existing reference graph.
    \item \textbf{Graph Update}: The agent updates existing node/edge attributes or infers graphs for newly encountered regions as it continues along its trajectory.
\end{itemize}

\subsubsection{\textbf{Graph Alignment}}
A partial query graph and a reference graph containing label information are given. When the query graph $G_q$ contains more than six nodes, an anchor node is randomly selected as the starting point for the search (line~\ref{line:12}). The process begins by identifying nodes in the reference graph that share the same label as the anchor node. For each candidate node, the search expands iteratively by traversing neighboring nodes in the query graph (line~\ref{line:13}) and attempting to find corresponding matches in the reference graph. This search recursively identifies triplet (node–edge–node) matches to maximize the alignment between the two graphs (line~\ref{line:1}-\ref{line:9}). The objective is to extract the intersection subgraph that best corresponds to the query graph. If the intersection subgraph exceeds the alignment threshold length \(\theta_{len}\), the corresponding nodes and edges are merged, and the 3D information of the reference graph is updated (line~\ref{line:15}-\ref{line:17}). Otherwise, if the match size falls below \(\theta_{len}\), the agent adds the query graph to the reference graph as new nodes and edges (line~\ref{line:18}).

\subsubsection{\textbf{Graph Update}}
If the partial query graph and the reference graph are aligned, the remaining triplets in the reference graph $G_r$ are updated in one of three ways, based on the newly recognized object \(O\) in $G_q$ to improve the accuracy of the 3D information:
\begin{enumerate}
\renewcommand\labelenumi{(\roman{enumi})}
    \item {Matching Node}: If the centroid distance between object \(O\) and an existing node \(v\) is below threshold \({\theta}_{dis}\), the Intersection over Union (IoU) between their bounding boxes exceeds \({\theta}_{bbox}\), and their labels are identical, then \(O\) and \(v\) are treated as the same. The 3D attributes of \(v\) are refined by updating its bounding box to the union of bounding boxes of \(O\) and \(v\), and the existing edge \(e\) is replaced with the corresponding edge from \(G_q\).

    \item {Conflicting Label}: If the centroid distance is less than \({\theta}_{dis}\) and the IoU exceeds \({\theta}_{bbox}\), but the labels differ, then the existing node \(v\) is replaced with a new node corresponding to \(O\). The new node inherits the class label and 3D spatial properties of \(O\), including the bounding box coordinates, and the edge \(e\) connected to \(v\) is also updated accordingly.

    \item {New Node}: If the centroid distance exceeds \({\theta}_{dis}\), \(O\) is regarded as novel and inserted into the scene graph as a new node. The corresponding edge structure from \(G_q\) is also added to maintain relational consistency.
\end{enumerate}

\algrenewcommand\algorithmicindent{0.8em}
\begin{algorithm}[!t]
\caption{Graph Alignment Algorithm for Agent $k$}
\label{alg:GA}
\small
\setlength{\baselineskip}{1.04\baselineskip}
\begin{algorithmic}[1]
\Statex \textbf{Input:} Query Graph $G_q$, Reference Graph $G_r$
\Statex \textbf{Output:} Updated 3D Scene Graph $G'_r$
\Statex

\Procedure{GraphSearch}{$q, G_q, G_r, map, visit$} \label{line:1}
    \State $visit \gets visit \cup \{q\}$
    \For{\textbf{each} neighbor $u$ of $q$ in $G_q$} \label{line:3}
        \If{$u \notin visit$}
            \For{\textbf{each} $v$ in neighbors of $map[q]$ in $G_r$}
                \If{$G_q[u].label = G_r[v].label$}
                    \State $map[u] \gets v$
                    \State $map \gets$ GraphSearch$(u, G_q, G_r, map, visit)$
                \EndIf
            \EndFor
        \EndIf
    \EndFor
    \State \Return $map$ \label{line:9}
\EndProcedure
\Statex

\Procedure{SceneGraphUpdate}{$G_q$, $G_r$}
    \State $map, visit \gets Init()$
    \State $anchors \gets SelectRandomNodes(G_q)$ \label{line:12} 
    \For{node $N_{anc}$ in $anchors$} \label{line:13}
        \State $map_{max}\!\gets$ GraphSearch$(N_{anc},\!G_q,\!G_r,\!map,\!visit)$
        \If{$len(map_{max}) >$ threshold $\;\theta_{len}$} \label{line:15}
            \State $G'_r \gets$ GraphUpdate$(G_q, G_r)$ \Comment{update $G_r$} \label{line:16}
            \State \Return $G'_r$ \label{line:17}
        \EndIf
    \EndFor
    \State $G'_r \gets G_r.Add(G_q)$ \Comment{just add $G_q$ as new graph} \label{line:18}
    \State \Return $G'_r$
\EndProcedure
\end{algorithmic}
\end{algorithm}

\section{MA3DSG-Bench}
\subsection{Evaluation Scenarios}
MA3DSG-Bench evaluates performance under two standard scenarios.
\subsubsection{\textbf{Static Collaborative Perception (SCP)}}
Multiple agents collaboratively operate within a large-scale indoor environment with no dynamic objects. Multiple agents simultaneously explore distinct regions and incrementally construct a unified 3D scene graph. This scenario evaluates the system’s capability to accurately and consistently integrate spatial and semantic information from diverse viewpoints into a cohesive graph representation.
\subsubsection{\textbf{Long-term Dynamic Collaborative Perception (LDCP)}}
This scenario introduces temporal dynamics, where changes occur over an extended period. Initially, agents construct a scene graph through collaborative exploration, similar to SCP. However, upon revisiting the same location after a significant time lapse, agents encounter changes such as objects that have moved, appeared, or disappeared entirely. Since such scenarios frequently occur in real-world indoor environments, LDCP assesses the system’s ability to detect and reconcile temporal inconsistencies by updating nodes and edges, thereby enhancing real-world applicability.

\subsection{Evaluation Dataset}
\begin{figure}[!t]
  \centering
  \includegraphics[width=\linewidth]{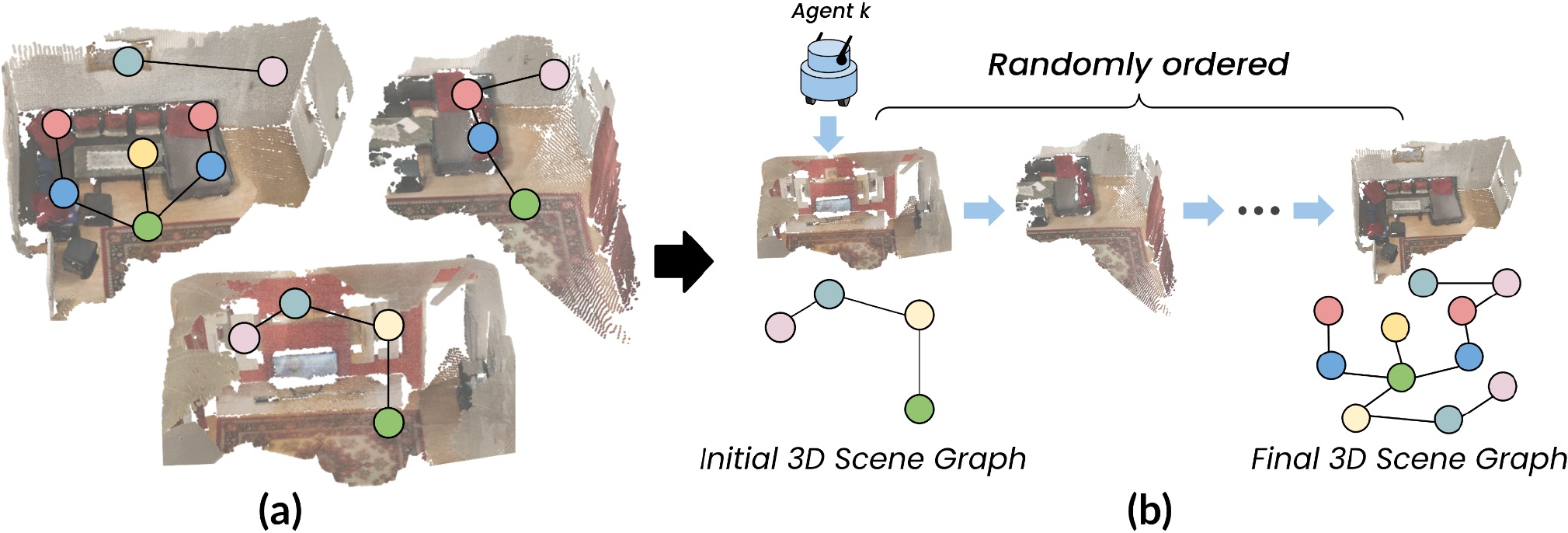}
  \caption{Unified domain evaluation. (a) Prior works treat each explored scene separately. (b) A newly annotated final 3D scene graph reflects temporal changes from randomly ordered visits for the LDCP scenario.}
  \label{fig:dataset}
\end{figure}

To evaluate the scalability of MA3DSG, we reformulate the 3RScan \cite{Wald2019RIO3O} and 3DSSG \cite{Wald2020Learning3S} datasets by assuming that all 47 room scenes in the test set can belong to \textit{a single, unified domain}, rather than evaluating them independently as in prior works. This unified setting, designed to evaluate robustness under diverse layouts, serves as a valid basis for assessing instance-level perception accuracy within rooms. As shown in Figure~\ref{fig:dataset}(a), although originating from the same physical environment, the scenes were previously treated as independent, with separate 3D scene graph generation and evaluation. Within the unified domain, as illustrated in Figure~\ref{fig:dataset}(b), agent $k$ incrementally constructs a 3D scene graph by exploring randomly ordered $\mathcal{S}^k$ in our benchmark. To simulate realistic multi-agent deployment, each agent’s trajectory is randomly generated while controlling the \textit{overlap ratio}---defined as the fraction of one agent’s trajectory intersecting another’s---to maintain a balanced and realistic coverage distribution across agents. In particular, we utilize the rescans of each room, where object configurations are rearranged, to generate a newly annotated final 3D scene graph that captures both spatial and temporal changes---including 478 dynamic objects, as shown in Table~\ref{tab:comparison6}.
\begin{table}[!t]
    \centering
    \setlength{\tabcolsep}{0.4em}
    \renewcommand{\arraystretch}{1}
    \caption{The number of Static/Dynamic objects.}
    \begin{tabular}{
        @{}
        >{\raggedright}m{1.9cm}|
        >{\centering\arraybackslash}m{1.3cm}
        *{3}{>{\centering\arraybackslash}m{1.3cm}}
    }
        \specialrule{0.9pt}{0pt}{1pt}
        \multirow{2}[1]{*}{\;\textbf{\# of Instances}} & \multirow{2}[1]{*}{\textbf{Static}} & \multicolumn{3}{c}{\textbf{Dynamic}} \\
        \cline{3-5}
        & & \textit{moved} & \textit{removed} & \textit{changed} \\
        \specialrule{0.8pt}{1pt}{1pt}
        \hspace{0.7cm}1,588 & 1,110 & 214 & 70 & 194 \\
        \specialrule{0.9pt}{1pt}{0pt}
    \end{tabular}
    \label{tab:comparison6}
    \vspace{1em}
    \caption{Unified 3D semantic scene graph sizes in the each setting.}
    \begin{tabular}{
        @{}
        >{\raggedright}m{2.2cm}
        >{\centering\arraybackslash}m{1.7cm}
        >{\centering\arraybackslash}m{1.7cm}
    }
        \specialrule{0.9pt}{0pt}{1pt}
        & \textbf{SCP} & \textbf{LDCP} \\
        \specialrule{0.8pt}{1pt}{1pt}
        \hspace{0.5cm}\textbf{\# of Nodes} & 1,588 & 1,518 \\
        \hspace{0.5cm}\textbf{\# of Edges} & 5,546 & 5,054 \\
        \specialrule{0.9pt}{1pt}{0pt}
    \end{tabular}
    \label{tab:graphsize}
\end{table}

\subsection{Metrics}
\subsubsection{Accuracy}
Following prior works \cite{Wu2021SceneGraphFusionI3, Wu2023Incremental3S}, we evaluate our method on triplet, object, and predicate prediction tasks. While prior work primarily reports recall@k, we also include precision and F1, as high recall alone does not ensure meaningful graph construction---excessive relationship predictions can introduce noise, reducing interpretability. Moreover, whereas previous studies focus only on semantic labels, we also incorporate spatial accuracy by considering the object’s center position and 3D bounding box IoU---ensuring alignment between the generated 3D scene graph and the environment’s actual geometry.
\subsubsection{Efficiency}
We assess the efficiency of the 3D scene graph generation system by computing the total graph alignment time, the whole scenario completion time, and the per-agent data traffic, defined as the mean total volume of data transmitted per agent.

\newcommand{\fcell}[1]{
    \begingroup
    \setbox\strutbox=\hbox{\vrule height 1.5ex depth 0.4ex width 0pt}
    \setlength{\fboxsep}{0pt}
    \colorbox{orange!40}{\makebox[0.75\linewidth][c]{\strut #1}}
    \endgroup}

\newcommand{\ffcell}[1]{
    \begingroup
    \setbox\strutbox=\hbox{\vrule height 1.5ex depth 0.4ex width 0pt}
    \setlength{\fboxsep}{0pt}
    \colorbox{orange!25}{\makebox[0.75\linewidth][c]{\strut #1}}
    \endgroup}

\newcommand{\fffcell}[1]{
    \begingroup
    \setbox\strutbox=\hbox{\vrule height 1.5ex depth 0.4ex width 0pt}
    \setlength{\fboxsep}{0pt}
    \colorbox{orange!10}{\makebox[0.75\linewidth][c]{\strut #1}}
    \endgroup}

\newcommand{\fcells}[3]{\fcell{#1} & \fcell{#2} & \fcell{#3}}
\newcommand{\ffcells}[3]{\ffcell{#1} & \ffcell{#2} & \ffcell{#3}}
\newcommand{\fffcells}[3]{\fffcell{#1} & \fffcell{#2} & \fffcell{#3}}

\begin{table*}[!t]
    \centering
    \caption{\textbf{Quantitative Evaluation of Accuracy and Efficiency under the SCP setting.}}
    \vspace{-2mm}
    \renewcommand{\arraystretch}{0.6} 
    \begin{tabular}{
        @{}>{\raggedright}m{2.5cm}
        @{}>{\centering\arraybackslash}m{2.5cm}
        @{\hskip 13pt}
        @{\extracolsep{-23pt}}*{3}{>{\centering\arraybackslash}m{1.4cm}}
        @{\hskip 13pt}
        @{\extracolsep{-23pt}}*{3}{>{\centering\arraybackslash}m{1.4cm}}
        @{\hskip 13pt}
        @{\extracolsep{-23pt}}*{3}{>{\centering\arraybackslash}m{1.4cm}}
        @{\extracolsep{1pt}}*{1}{>{\centering\arraybackslash}m{1.1cm}}
        @{\extracolsep{1pt}}*{1}{>{\centering\arraybackslash}m{1.1cm}}
        @{\extracolsep{1pt}}*{1}{>{\centering\arraybackslash}m{1.1cm}}@{}}
            \toprule 
            \multirow{3}{*}{\quad\quad\quad\textbf{Method}}
            & \multirow{3}{*}{\textbf{\makecell{Domain Size \\ (\# of rooms)}}}
            & \multicolumn{3}{c}{\makecell{\textbf{Triplet}}}
            & \multicolumn{3}{c}{\textbf{Object}}
            & \multicolumn{3}{c}{\makecell{\textbf{Predicate}}}
            & \multirow{3}{*}{\textbf{\makecell{Traffic. \\ (MB)}}}
            & \multirow{3}{*}{\textbf{\makecell{Align. \\ (sec)}}}
            & \multirow{3}{*}{\textbf{\makecell{Total. \\ (min)}}} \\
            \cmidrule(l{1.3em}r{2.7em}){3-5} \cmidrule(l{1.3em}r{2.7em}){6-8} \cmidrule(l{1.3em}r{1.2em}){9-11}
               & & R@1 & P@1 & F1@1 & R@1 & P@1 & F1@1 & R@1 & P@1 & F1@1 \\
            \midrule
            \;\underline{\textit{\textbf{\textcolor{blue!85!white}{Single-agent approach}}}} & \\
             & 5 & 3.1 & 3.2 & 3.1 & 27.1 & 32.5 & 29.5 & 15.2 & 19.5 & 17.0 & - & - & -\\
             \multirow{3}{*}{\;3DSSG} 
             & 15 & \fcells{16.8}{15.5}{16.1} & 31.5 & 37.7 & 34.4 & \fcells{26.3}{39.2}{31.5} & - & - & -\\
             & 25 & \fcells{18.6}{15.5}{16.9} & 32.6 & 36.5 & 34.4 & \fcells{29.7}{37.9}{33.3} & -& - & -\\
             & 47 & 14.8 & 9.0 & 11.2 & 34.0 & 33.4 & 33.7 & \fffcells{25.9}{26.8}{26.3}  & - & - & -\\
             \midrule
             & 5 & \ffcells{19.6}{7.6}{10.9} & \fffcells{52.1}{33.1}{40.5} & \ffcells{25.9}{23.4}{24.6}  & - & - & 6.4 \\
              \multirow{3}{*}{\;SGFN} 
             & 15 & \ffcells{20.4}{9.5}{13.0} & \fcells{51.7}{30.7}{38.5} & \ffcells{25.7}{26.3}{26.0} & - & - & 17.1 \\
             & 25 & \fffcells{24.3}{10.5}{14.7} & \fcells{55.8}{30.7}{39.6} & \fffcells{30.5}{27.7}{29.0} & - & - & 32.7\\
             & 47 & \fcells{26.4}{9.6}{14.1} & \fcells{56.2}{29.4}{38.6} & \fcells{32.0}{25.8}{28.6} & - & - & 61.8\\
            \midrule
            \;\underline{\textit{\textbf{\textcolor{green!65!black}{Multi-agent approach}}}} & \\
             & 5 & \fffcells{14.7}{5.4}{7.9} & \ffcell{50.0} & \ffcell{34.3} & \ffcell{40.7} & 18.9 & 16.4 & 17.6 & 43.2 & 11.7 & 2.5  \\
              \multirow{3}{*}{\;SGFN + SGAligner} 
             & 15 & 10.8 & 4.7 & 6.6 & \fffcells{46.1}{30.5}{36.7} & 14.6 & 15.8 & 15.2 & 116.2 & 35.1 & 4.9  \\
             & 25 & 16.2 & 6.9 & 9.7 & \ffcell{51.1} & \ffcell{31.7} & \ffcell{39.1} & 20.9 & 20.2 & 20.5 & 181.7 & 56.5 & 9.1 \\
             & 47 & \fffcells{22.5}{8.8}{12.7} & \ffcell{50.0} & \ffcell{31.3} & \ffcell{38.5} & 27.5 & 24.6 & 25.9 & 364.2 & 107.1 & 16.6 \\
             \midrule
             & 5 & 13.8 & 5.2 & 7.5 & \fcell{52.1} & \fcell{33.8} & \fcell{41.0} & \fffcells{19.6}{17.3}{18.4} & 43.1 & \underline{3.50} & \underline{2.4}  \\
              \multirow{3}{*}{\;SGFN + SG-PGM} 
             & 15 & 10.6 & 4.6 & 6.5 & \ffcell{46.7} & \ffcell{30.4} & \ffcell{36.8} & 14.8 & 16.0 & 15.4 & 116.1 & \underline{10.5} & \underline{4.5}  \\
             & 25 & 16.2 & 6.9 & 9.7 & \fffcells{51.3}{31.0}{38.6} & 21.2 & 20.4 & 20.8 & 181.6 & \underline{16.9} & \underline{8.5}  \\
             & 47 & 22.5 & 8.8 & 12.6 & \fffcells{50.0}{31.0}{38.3} & \ffcells{28.1}{25.4}{26.7} & 364.1 & \underline{32.1} & \underline{15.3} \\
             \midrule
             & 5 & \fcell{21.1} & \fcell{7.9} & \fcell{11.5} & 47.9 & 30.7 & 37.4 & \fcells{25.9}{24.0}{24.9} & \textbf{0.3} & \textbf{0.0} & \textbf{2.3} \\
              \multirow{3}{*}{\;\textbf{MA3DSG (Ours)}} 
             & 15 & \fffcells{17.3}{8.9}{11.7} & 51.4 & 25.5 & 34.1 & \fffcells{21.1}{22.1}{21.6} & \textbf{1.0} & \textbf{0.01} & \textbf{4.3} \\
             & 25 & \ffcell{25.7} & \ffcell{11.8} & \ffcell{16.2} & 54.9 & 27.2 & 36.4 & \ffcells{31.2}{28.5}{29.8} & \textbf{1.7} & \textbf{0.01} & \textbf{8.2} \\
             & 47 & \ffcell{24.2} & \ffcell{9.5} & \ffcell{13.7} & 55.8 & 25.6 & 35.1 & 27.7 & 22.2 & 24.6 & \textbf{3.7} & \textbf{0.02} & \textbf{14.8} \\
            \bottomrule
            \vspace{1mm}
            \makebox[\textwidth][l]{ 
            \hspace{-2mm} \parbox{0.99\textwidth}{
                \scriptsize * For each domain size, the top three F1 scores for each metric group are highlighted using three levels of color intensity, while the best efficiency result is shown in bold.}}
    \end{tabular}
    \label{tab:SCP}
    \vspace{-2mm}
\end{table*}

\begin{table*}[!ht]
    \centering
    \caption{\textbf{Quantitative Evaluation of Accuracy and Efficiency under the LDCP setting.}}
    \vspace{-2mm}
    \renewcommand{\arraystretch}{0.6} 
    \begin{tabular}{
        @{}>{\raggedright}m{2.5cm}
        @{}>{\centering\arraybackslash}m{2.5cm}
        @{\hskip 13pt}
        @{\extracolsep{-23pt}}*{3}{>{\centering\arraybackslash}m{1.4cm}}
        @{\hskip 13pt}
        @{\extracolsep{-23pt}}*{3}{>{\centering\arraybackslash}m{1.4cm}}
        @{\hskip 13pt}
        @{\extracolsep{-23pt}}*{3}{>{\centering\arraybackslash}m{1.4cm}}
        @{\extracolsep{1pt}}*{1}{>{\centering\arraybackslash}m{1.1cm}}
        @{\extracolsep{1pt}}*{1}{>{\centering\arraybackslash}m{1.1cm}}
        @{\extracolsep{1pt}}*{1}{>{\centering\arraybackslash}m{1.1cm}}@{}}
            \toprule 
            \multirow{3}{*}{\quad\quad\quad\textbf{Method}}
            & \multirow{3}{*}{\textbf{\makecell{Domain Size \\ (\# of rooms)}}}
            & \multicolumn{3}{c}{\makecell{\textbf{Triplet}}}
            & \multicolumn{3}{c}{\textbf{Object}}
            & \multicolumn{3}{c}{\makecell{\textbf{Predicate}}}
            & \multirow{3}{*}{\textbf{\makecell{Traffic. \\ (MB)}}}
            & \multirow{3}{*}{\textbf{\makecell{Align. \\ (sec)}}}
            & \multirow{3}{*}{\textbf{\makecell{Total. \\ (min)}}} \\
            \cmidrule(l{1.3em}r{2.7em}){3-5} \cmidrule(l{1.3em}r{2.7em}){6-8} \cmidrule(l{1.3em}r{1.2em}){9-11}
                & & R@1 & P@1 & F1@1 & R@1 & P@1 & F1@1 & R@1 & P@1 & F1@1 \\
            \midrule
            \;\underline{\textit{\textbf{\textcolor{blue!85!white}{Single-agent approach}}}} & \\
             & 5 & \ffcells{3.1}{3.2}{3.1} & 24.0 & 28.7 & 26.1 & \ffcells{12.2}{15.8}{13.8} & - & - & -\\
             \multirow{3}{*}{\;3DSSG} 
             & 15 & \fcells{11.2}{9.4}{10.2} & 27.7 & 32.8 & 30.1 & \fcells{18.3}{26.6}{21.7} & - & - & -\\
             & 25 & \fcells{13.3}{9.9}{11.4} & 28.1 & 30.8 & 29.4 & \fcells{22.1}{26.8}{24.2} & - & - & -\\
             & 47 & \fcells{10.9}{6.0}{7.7} & 29.5 & 28.6 & 29.1 & \fcells{20.6}{20.3}{20.5} &- & - & -\\
             \midrule
             & 5 & 4.2 & 1.4 & 2.1 & 44.8 & 28.5 & 24.8 & 9.2 & 8.2 & 8.7 & - & - & 22.2 \\
             \multirow{3}{*}{\;SGFN} 
             & 15 & 9.2 & 3.6 & 5.2 & \ffcells{44.6}{26.2}{33.0} & 14.2 & 13.8 & 14.0 & - & - & 69.7 \\
             & 25 & \fffcells{12.1}{4.3}{6.3} & \ffcells{47.9}{25.8}{33.5} & 18.3 & 15.3 & 16.7 & - & - & 95.8 \\
             & 47 & \ffcells{14.2}{4.3}{6.6} & \ffcells{47.6}{24.6}{32.4} & 19.5 & 14.6 & 16.7 & - & - &166.7 \\
            \midrule
            \;\underline{\textit{\textbf{\textcolor{green!65!black}{Multi-agent approach}}}} & \\
             & 5 & \fffcells{6.1}{1.9}{2.9} & \ffcells{44.8}{29.3}{35.4} & \fffcells{14.6}{12.5}{13.5} & 33.1 &7.1 & 124.9\\
              \multirow{3}{*}{\;SGFN + SGAligner} 
             & 15 & \fffcells{8.9}{3.9}{5.4} & 36.9 & 27.3 & 31.4 & \fffcells{17.5}{20.7}{18.9} &  408.1 & 124.7 &18.5  \\
             & 25 & 8.8 & 3.3 & 4.8 & 41.6 & 25.8 & 31.8 & 17.5 & 16.1 & 16.8 & 575.6 & 185.1 & 25.0  \\
             & 47 & \fffcells{11.9}{4.3}{6.4} & \fffcells{39.6}{25.4}{31.0} & \ffcells{21.2}{18.8}{19.9} & 1013.2 & 350.6 & 46.8 \\
             \midrule
             & 5 & 5.1 & 1.6 & 2.4 & \fcells{46.9}{29.0}{35.9} & \fcells{15.4}{13.4}{14.3} & 124.8 & \underline{9.91} & \underline{6.7} \\
              \multirow{3}{*}{\;SGFN + SG-PGM} 
             & 15 & 8.6 & 3.8 & 5.3 & \fffcells{37.6}{27.3}{31.6} & \ffcells{17.7}{21.0}{19.2} & 408.0 & \underline{37.3} & \underline{17.0} \\
             & 25 & 8.6 & 3.3 & 4.7 & \fffcells{42.0}{25.8}{31.9} & \ffcells{17.7}{16.3}{17.0} & 575.5 & \underline{55.4} & \underline{22.7} \\
             & 47 & 11.1 & 4.1 & 6.0 & 39.4 & 25.3 & 30.8 & \fffcells{20.1}{17.9}{19.0} & 1013.1 & \underline{104.9} & \underline{42.7} \\
             \midrule
             & 5 & \fcells{8.1}{2.6}{4.0} & \fffcells{43.8}{28.4}{34.4} & 11.0 & 9.6 & 10.2 & \textbf{1.0} & \textbf{0.03} &\textbf{6.5} \\
              \multirow{3}{*}{\;\textbf{MA3DSG (Ours)}} 
             & 15 & \ffcells{10.6}{4.2}{6.0} & \fcells{44.9}{26.4}{33.3} & 16.7 & 16.2 & 16.4 & \textbf{4.5} & \textbf{0.13} &\textbf{16.4}  \\
             & 25 & \ffcells{13.4}{4.7}{7.0} & \fcells{49.1}{25.8}{33.9} & \fffcells{18.8}{15.4}{16.9} & \textbf{6.1} & \textbf{0.19} &\textbf{21.8} \\
             & 47 & 13.1 & 4.0 & 6.2 & \fcells{48.3}{25.0}{33.0} & 18.9 & 14.1 & 16.2 & \textbf{11.6} & \textbf{0.37} &\textbf{41.0} \\
            \bottomrule
            \vspace{1mm}
            \makebox[\textwidth][l]{ 
            \hspace{-2mm} \parbox{0.99\textwidth}{
                \scriptsize * For each domain size, the top three F1 scores for each metric group are highlighted using three levels of color intensity, while the best efficiency result is shown in bold.} }
    \end{tabular}
    \label{tab:LDCP}
    \vspace{-5mm}
\end{table*}
\begin{figure*}[!t]
  \centering
  \includegraphics[width=\linewidth]{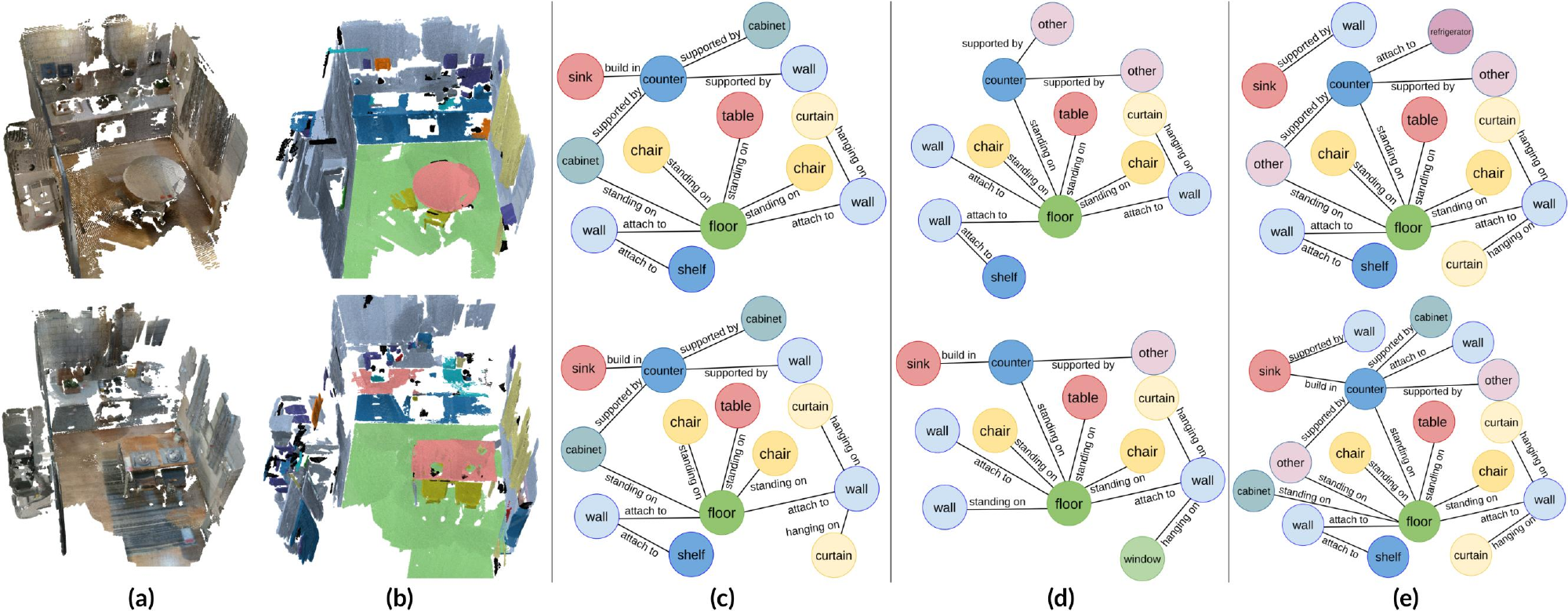}
  \vspace{-7mm}
  \caption{Qualitative results of SGFN and MA3DSG. We visualize (a) incrementally scanned point clouds, (b) ground truth instance segmentation, (c) ground truth 3D Semantic Scene Graph, (d) SGFN-generated, and (e) MA3DSG-generated 3D Semantic Scene Graphs. For the same room, the upper row shows SCP results and the lower row shows LDCP results.}
  \label{fig:quality}
  \vspace{-4mm}
\end{figure*}

\section{Experiments}
\subsection{Baselines}
We compare our \textbf{MA3DSG} with two types of baselines:
\begin{itemize}
    \item \textbf{Single-Agent Baselines}: Conventional 3DSGG research includes 3DSSG \cite{Wald2020Learning3S} and SGFN \cite{Wu2021SceneGraphFusionI3}—the prevalent setting. In these methods, a single agent is responsible for generating the complete 3D scene graph. Unlike other baselines, the 3DSSG processes the entire point cloud at once, making total runtime measurement inappropriate for evaluating incremental 3D scene graph generation.
    \item \textbf{Multi-Agent Baseline}: Due to the challenges of constructing multi-agent systems, the literature lacks established baselines for direct comparison. To address this, we introduce strong baselines by combining SGFN with recent modules: SGAligner \cite{sarkar2023sgaligner} and SG-PGM \cite{Xie2024SGPGMPG}, denoted as \textit{SGFN+SGAligner} and \textit{SGFN+SG-PGM}.
\end{itemize}

\subsection{Comparative Studies}
In all experiments, we set five agents with an overlap ratio of 0.2 for the multi-agent setups. For MA3DSG, several thresholds were determined empirically through ablations, where we set $\theta_{dis}$ to 1.5 meters, $\theta_{len}$ to 3, and $\theta_{bbox}$ to 0.4.
\subsubsection{\textbf{Static Collaborative Perception (SCP)}} Table~\ref{tab:SCP} compares MA3DSG against single-agent and multi-agent baselines in the SCP setting, demonstrating that it achieves performance comparable to both. Across all domain sizes, MA3DSG maintains R@1, P@1, and F1@1 on par with those of the single-agent SGFN, with no substantial decline. Specifically, MA3DSG exhibits deviations in triplet F1@1 ranging from +1.5/-1.3\%, in object F1@1 from -3.1/-4.4\%, and in predicate F1@1 from +0.8/-4.4\%. MA3DSG also operates \textbf{\textit{2.8$\times$ to 4.2$\times$ faster}} than SGFN, with the advantage increasing as domain size scales. This stability stems from MA3DSG’s incremental updates, which preserve prior information and leverage multi-agent observations to address unscanned regions---unlike full graph replacement in single-agent approaches. However, this mechanism introduces potential errors, such as conservative predictions from inconsistent updates, missed relationships, or noise accumulation over iterative updates, especially in large domains. In contrast, SGFN+SG-PGM exhibits a more pronounced performance gap relative to SGFN with deviations in Triplet F1@1 ranging from -1.5/-6.5\%, in object F1@1 from +0.5/-1.7\%, and in predicate F1@1 from -1.9/-10.6\%.

\subsubsection{\textbf{Long-term Dynamic Collaborative Perception (LDCP)}} Table~\ref{tab:LDCP} shows quantitative results for the LDCP setting, which introduces temporal inconsistencies requiring adaptive refinement and continuous 3D scene graph updates. When comparing SGFN and MA3DSG, the deviations range up to +1.9/-0.4\% in triplet F1@1, +9.6/+0.3\% in object F1@1, and +2.4/-0.5\% in predicate F1@1. MA3DSG operated \textbf{\textit{3.4$\times$ to 4.1$\times$ faster}} in terms of processing efficiency, with greater gains as domain size increased. These results underscore MA3DSG’s ability to adeptly manage dynamic scene changes by integrating multi-agent observations and refining graphs incrementally. In contrast, the comparison between SGFN+SG-PGM and SGFN shows F1@1 deviations ranging from +0.3/-1.6\% in triplet, +11.1/-1.6\% in object, and from +5.6/+0.3\% in predicate. While SGFN+SG-PGM leverages multi-agent collaboration to achieve faster execution than SGFN, its dependency on point cloud registration leads to higher processing time and computational cost than MA3DSG.

\subsection{Efficiency Analysis}
\subsubsection{Alignment Time}
MA3DSG substantially reduces computational overhead compared to multi-agent baselines. It achieves consistently low alignment latency—0.02 seconds in SCP and 0.37 seconds in LDCP—even in large-scale domains. In contrast, SGFN+SG-PGM incurs significantly higher alignment costs, highlighting the efficiency of our multi-agent communication and scene graph integration. Notably, MA3DSG performs inference entirely on CPU, emphasizing its lightweight and hardware-efficient design.

\subsubsection{Data Traffic}
To assess the communication efficiency, we compute traffic differently depending on the agent configuration. For the multi-agent baselines, the communication cost includes both the graph data and the point cloud data exchanged among agents. In contrast, MA3DSG transmits only lightweight graph representations rather than full point cloud, resulting in substantially lower traffic overhead. MA3DSG reduces communication cost by a factor of \textit{98.4$\times$} in the SCP and \textit{87.3$\times$} in the LDCP---highlighting its strong scalability in extremely large-scale 3D environments.

\subsection{Qualitative Results}
Figure~\ref{fig:quality} presents a qualitative comparison of 3D scene graphs under realistic, dynamic indoor environments, projected onto 2D for visualization purposes. The top row corresponds to the initial reference scan, while the bottom row shows the rescan data acquired after a time interval when an agent revisits the same room. As observed in (a) and (b), notable scene changes occur; the table changes from circular to rectangular, various furniture shifts, and lighting conditions vary due to open curtains. Notably, MA3DSG merges prior graphs when later agents encounter unscanned areas, preserving richer nodes such as shelves and cabinets, as well as edges that SGFN often overlooks.

\section{Conclusion}
We introduced MA3DSG, a multi-agent framework for 3D scene graph generation that incrementally updates graphs, leveraging shared agent knowledge to achieve scalability and efficiency in large-scale settings. In the process, we developed MA3DSG-Bench, a novel benchmark tailored to evaluate 3DSGG scalability across diverse agent configurations, domain sizes, and dynamic conditions---surpassing prior single-agent, small-scale benchmarks. Together, these contributions establish a robust foundation for multi-agent 3DSGG research. Future work will refine the graph update mechanism, enhance robustness to scene variations, and optimize computational performance in dynamic environments. These efforts aim to advance scalable multi-agent 3DSGG systems, setting a new standard for the field.





\section*{ACKNOWLEDGMENT}
{\footnotesize
    This research was partly supported by the Institute of Information \& Communications Technology Planning \& Evaluation (IITP) grant funded by the Korea government (MSIT) (No. RS-2022-II220907); by the National Research Foundation of Korea (NRF) grant funded by the Korea government (MSIT) (No. NRF-2022R1C1C1009989); and by the National Research Council of Science \& Technology (NST) grant funded by the Korea government (MSIT) (No. GTL25041-000).
}


\bibliographystyle{ICRA/IEEEtran} 
\bibliography{ICRA/IEEEabrv,ICRA/root}

\end{document}